\documentclass[runningheads]{llncs}
\usepackage[T1]{fontenc}
\usepackage{graphicx}
\usepackage{cite}
\usepackage{amsmath,amssymb,amsfonts}
\usepackage{textcomp}
\usepackage{xcolor}
\usepackage{subcaption}

\usepackage{algorithm}
\usepackage{algpseudocode}

\usepackage[numbers]{natbib}

\begin{document}
\title{Self-supervised Conformal Prediction for Uncertainty Quantification in Imaging Problems}
\titlerunning{SURE-based Conformal Prediction for UQ in Imaging}
\author{Jasper M. Everink\inst{1}\orcidID{0000-0001-7263-0317} \and Bernardin Tamo Amougou\inst{2,3} \and Marcelo Pereyra\inst{2}\orcidID{0000-0001-6438-6772} }
\authorrunning{J.M. Everink, B. Tamo Amougou and  M. Pereyra.}
\institute{Technical University of Denmark, Kgs. Lyngby, Denmark, \email{jmev@dtu.dk} \and Heriot-Watt University \& Maxwell Institute for Mathematical Sciences, Edinburgh, UK \and Universit\'e de Paris Cit\'e, Paris, France }

\maketitle              % typeset the header of the contribution
\begin{abstract}
Most image restoration problems are ill-conditioned or ill-posed and hence involve significant uncertainty. Quantifying this uncertainty is crucial for reliably interpreting experimental results, particularly when reconstructed images inform critical decisions and science. However, most existing image restoration methods either fail to quantify uncertainty or provide estimates that are highly inaccurate. Conformal prediction has recently emerged as a flexible framework to equip any estimator with uncertainty quantification capabilities that, by construction, have nearly exact marginal coverage. To achieve this, conformal prediction relies on abundant ground truth data for calibration. However, in image restoration problems, reliable ground truth data is often expensive or not possible to acquire. Also, reliance on ground truth data can introduce large biases in situations of distribution shift between calibration and deployment. This paper seeks to develop a more robust approach to conformal prediction for image restoration problems by proposing a self-supervised conformal prediction method that leverages Stein's Unbiased Risk Estimator (SURE) to self-calibrate itself directly from the observed noisy measurements, bypassing the need for ground truth. The method is suitable for any linear imaging inverse problem that is ill-conditioned, and it is especially powerful when used with modern self-supervised image restoration techniques that can also be trained directly from measurement data. The proposed approach is demonstrated through numerical experiments on image denoising and deblurring, where it delivers results that are remarkably accurate and comparable to those obtained by supervised conformal prediction with ground truth data.
%Most image restoration problems are not well-posed, and thus may have some significant intrinsic uncertainty. Robustly quantifying the uncertainty in the solutions to such problems is important for the reliable interpretation of experimental results, especially if the reconstructed images are used as evidence in decision-making or science. Unfortunately, most image restoration methods do not quantify the uncertainty in the restored images, or provide uncertainty quantification estimates that are highly subjective, inaccurate, and not useful in practice yet. Conformal prediction has recently emerged as a powerful framework to endow any statistical estimator with uncertainty quantification capabilities. Conformal prediction is versatile and easy to apply to problems of small or moderate dimension. However, when applied to imaging problems and other high-dimensional problems, it often provides uncertainty estimates that are too loose and uninformative. In this paper, we propose a conformal prediction method tailored for linear imaging inverse problems. The proposed method leverages Stein's unbiased risk estimator to sharpen its uncertainty quantification results and scale robustly to high-dimensional settings. The approach is demonstrated through a series of numerical experiments related to image denoising and deblurring, where it is used to construct confidence regions for a wide range of different imaging methods.

\end{abstract}

\section{Introduction}
Image restoration tasks often carry a significant amount of uncertainty, as they admit a wide variety of solutions that are equally in agreement with the observed data. Quantifying and characterizing this uncertainty is important for applications that depend on restored images to inform critical decisions. Several statistical frameworks exist to address uncertainty quantification (UQ) in imaging sciences. Among these, the Bayesian statistical framework has been extensively studied and applied \cite{JariErkki}, enabling the incorporation of prior knowledge and providing probabilistic interpretations of uncertainty. This framework supports a diverse range of modeling and algorithmic approaches, as demonstrated in works such as \cite{Durmus2018,Pereyra2017,laumont2022bayesian,holden2022bayesian}. Unfortunately, despite significant progress in the field, state-of-the-art Bayesian imaging methods are not yet able to provide accurate UQ on structures that are larger than a few pixels in size \cite{thong2024bayesianimagingmethodsreport}. 

With regards to non-Bayesian approaches to UQ in image restoration, the recently proposed equivariant bootstrapping method \cite{tachella2023equivariant} offers excellent frequentist accuracy, even for large image structures. This is achieved by exploiting known symmetries in the problem to reduce the bias inherent to bootstrapping. Equivariant bootstrapping is particularly accurate in problems that are highly ill-posed, such as compressive sensing, inpainting and limited angle tomography and radio-interferometry, as in these cases the bias from bootstrapping is almost fully removed by the actions of the symmetry group \cite{tachella2023equivariant,Liaudat2024}. Conversely, equivariant bootstrapping struggles with problems that are not ill-posed, such as image denoising or mild deblurring, as in this case it is difficult to identify symmetries to remove the bias from bootstrapping (see \cite{tachella2023equivariant} for details).

Moreover, when sufficient ground truth data are available for calibration, conformal prediction presents another highly flexible strategy for UQ in image restoration. Crucially, conformal prediction can be seamlessly applied to any image restoration technique to deliver UQ results that, by construction, have nearly exact marginal coverage \cite{angelopoulos2021gentle}. Also, conformal prediction can be easily combined with other UQ strategies, such as Bayesian imaging strategies (see, e.g., \cite{narnhofer2024posterior}) or equivariant bootstrapping \cite{Liaudat2024}, as a correction step. %As a result, there is significant interest in incorporation conformal prediction within imaging pipelines.

However, as mentioned previously, in its original form, conformal prediction approaches require access to abundant ground truth data for calibration. In many applications, access to ground truth data is either expensive or impossible. Also, reliance on ground truth data for calibration can lead to poor accuracy in situations of distribution shift between calibration and deployment. To address this limitation of conformal prediction, we propose a self-supervised conformal prediction method that leverages Stein's Unbiased Risk Estimator (SURE) to self-calibrate UQ results directly from the observed noisy measurements, bypassing the need for ground truth.

The remainder of this paper is organized as follows. Section \ref{sec:background} provides an overview of conformal prediction and a formal problem statement. Section \ref{sec:method} introduces the proposed self-supervised conformal prediction method. Section \ref{sec:experiments} demonstrates the proposed approach through numerical experiments on image denoising and non-blind deblurring tasks and by considering model-based as well as learning-based estimators. Conclusions and perspectives for future work are finally reported in Section \ref{sec:discussion}.

\section{Problem statement}\label{sec:background}

We consider the estimation of a set of plausible values for an unknown image of interest \( x^\star \in \mathbb{R}^n \), from a noisy degraded measurement \( y \in \mathbb{R}^m \). We assume that \( x^\star \) is a realization of a random variable \( X \), and \( y \) is a realization of the conditional random variable \( Y | X = x^\star \). A point estimator for \( x^\star \), derived from \( Y \), is henceforth denoted by \( \hat{x}(Y) \). We focus on the case where observations follow a Gaussian noise model:
\[
(Y | X = x^\star) \sim \mathcal{N}(Ax^\star, \sigma^2 \mathbb{I}_m),
\]
where \( A \in \mathbb{R}^{m \times n} \) models deterministic instrumental aspects of the image restoration problem, \( \sigma^2 > 0 \) is the noise variance, and \( \mathbb{I}_m \) is the \( m \times m \) identity matrix. Throughout the paper we assume $A$ is a full-rank but potentially highly ill-conditioned linear operator. %The generalisation to other noise models, such as Poisson, and Poisson-Gaussian is straightforward and will be discussed later. 

%The observation \( Y \) represents a degraded version of the underlying true signal \( x^\star \), projected into a lower-dimensional measurement space and corrupted by additive Gaussian noise.

Our goal is to construct a region \( C(Y) \subset \mathbb{R}^n \) in the solution space such that

\begin{equation}\label{predictionC}
    \mathbb{P}_{(X, Y)}\left(X \in C(Y) \right) \geq 1-\alpha\,,
\end{equation}
where the probability is taken with respect to the joint distribution of \( (X, Y) \), and \( \alpha \in [0,1] \) specifies the desired confidence level.

To illustrate, suppose that \( x^\star \) is a high-resolution MRI scan of an adult brain. The random variable \( X \) characterizes the distribution of brain MRI scans for a generic individual within the population, as obtained by an ideal noise-free and resolution-perfect MRI scanner. The specific image \( x^\star \) corresponds to an MRI scan of a particular individual, while \( y \) represents the noisy, degraded measurement acquired in practice. The estimator \( \hat{x}(y) \) produces an estimate of \( x^\star \). The region \( C(y) \) encapsulates a set of plausible solutions, rather than a single estimate, and satisfies the guarantee in \eqref{predictionC}. This means that if the procedure is repeated across a large number of individuals from the population, the constructed regions \( C \) will contain the respective true images \( x^\star \) in at least \( 1-\alpha\) of the cases.

Conformal prediction provides a general framework for constructing sets \( C \) with the desired probabilistic guarantee \cite{angelopoulos2021gentle}. This is achieved using a \emph{non-conformity measure} \( s: \mathbb{R}^n \times \mathbb{R}^m \to \mathbb{R} \). By computing the top \((1-\alpha)\)-quantile \( q_\alpha \) of the statistic \( s(X, Y) \), the set \( C(y) \) is defined as:
\[
C(y) := \{x \in \mathbb{R}^n \,|\, s(x, y) \leq q_\alpha\} \quad \text{for all } y \in \mathbb{R}^m.
\]
By construction, this set satisfies:
\[
\mathbb{P}_{(X, Y)} \big(X \in C(Y)\big) = \mathbb{P}_{(X, Y)} \big(s(X, Y) \leq q_\alpha \big) \geq 1 - \alpha,
\]
for any confidence level \( \alpha \in (0, 1) \). With a sufficiently large sample \(\{x_i, y_i\}_{i=1}^M\) to calibrate \( q_\alpha \), any suitable function \( s \) can be used to construct a set \( C \) that contains \( x^\star \) with high probability under the joint distribution of \((X, Y)\).

A popular specific implementation of this framework is \emph{split conformal prediction}. Given a training sample \(\{X_i, Y_i\}_{i=1}^M\) of independent (or exchangeable) realizations of \((X, Y)\), the method estimates the top \(\frac{\lceil(M+1)(1-\alpha)\rceil}{M}\)-quantile \(\hat{Q}_\alpha\) of \(\{s(X_i, Y_i)\}_{i=1}^M\). For a new observation \((X_{\text{new}}, Y_{\text{new}})\), the prediction set is then:
\[
\hat{C}(Y_{\text{new}}) := \{X_{\text{new}} \in \mathbb{R}^n \,|\, s(X_{\text{new}}, Y_{\text{new}}) \leq \hat{Q}_\alpha\}.
\]
This set satisfies the guarantee:
\begin{equation}\label{eq:conformal_guarantee}
    \mathbb{P}_{(X, Y)^{M+1}}\left(X_{\text{new}} \in \hat{C}(Y_{\text{new}}) \right)  
    \geq 1-\alpha\,,
\end{equation}
where the probability accounts for the joint distribution of the \( M \) training samples and the new observation. Notably, this formulation includes a finite-sample correction because \(\hat{Q}_\alpha\) is derived from the \(\frac{\lceil(M+1)(1-\alpha)\rceil}{M}\)-quantile. This correction becomes negligible as \( M \to \infty \).
For an excellent introduction to conformal prediction, please see \cite{angelopoulos2021gentle}.

While the conformal prediction framework is highly flexible, not all non-conformity measures \( s(x, y) \) deliver prediction regions that are equally useful in practice. Indeed, all prediction sets \( \hat{C}(y) \) take the form of a sub-level set of the function \( x \mapsto s(x, y) \), which can be arbitrarily chosen. As a result, it is possible to construct infinitely many regions in \( \mathbb{R}^n \) that satisfy the guarantee of containing the true solution \( x^\star \) with probability at least \( 1-\alpha \). However, many of these regions may be excessively large, especially in high-dimensional settings, where poorly designed score functions \( s(x, y) \) can lead to regions \( \hat{C}(y) \) that are overly conservative and cover most of the support of \( X \). 

Carefully designing \( s(x, y) \) allows obtaining conformal prediction sets that are compact and informative, even in high dimension. In particular, it is essential that \( s(x, y) \) is constructed in a way that reduces the variability of \( s(X, Y) \)% and allows scaling to large problems without Several approaches have been recently proposed to scale conformal prediction to high-dimensional problems (see, e.g., \cite{johnstone2021conformal, messoudi2020conformal, messoudi2021copula, messoudi2022ellipsoidal})
. Of particular interest are normalized non-conformity measures of the form \cite{johnstone2021conformal}:
\begin{equation}\label{eq:score_multi_target}
s(x, y) = \|x - \hat{x}(y)\|^2_{\Sigma(y)} = \left(x - \hat{x}(y)\right)^\top \Sigma(y) \left(x - \hat{x}(y)\right),
\end{equation}
where \( \Sigma(y) \) is a positive definite matrix of size \( n \times n \). A well-chosen \( \Sigma(y) \) reduces variability in \( s(X, Y) \), leading to prediction sets that are well-centered around \( \hat{x}(y) \), compact, and highly informative. In practice, \( \Sigma(y) \) is often chosen as an approximation of the inverse-covariance matrix of the error $X-\hat{x}(Y)$ \cite{johnstone2021conformal}.

%These sets satisfy property \eqref{eq:conformal_guarantee} by construction.

%In prior work, \( \Sigma(y) \) is often chosen as an approximate inverse-covariance matrix. For instance, \cite{johnstone2021conformal} propose estimating this matrix globally based on the errors in a sample \( \{x_i, y_i\}_{i=1}^M \). However, this global approach neglects local variations in \( y \), which are essential in high-dimensional problems. To address this, \cite{messoudi2022ellipsoidal} suggest refining the global matrix by incorporating errors from the \( k \)-nearest neighbors \( \{x_i, y_i\}_{i=1}^k \) of \( y \). While this method captures some local variability, it requires a large amount of data to accurately estimate the local uncertainty.

A fundamental obstacle to applying conformal prediction to image restoration problems is the need for paired samples \( \{x_i, y_i\} \), as obtaining the ground truth image \( x_i \) is precisely the goal of solving the imaging problem in the first place. This can be partially mitigated by relying on a training dataset, at the risk of delivering poor results in situations of distribution shift (e.g., returning to our illustrative example related to MRI imaging, when the population encountered in deployment differs significantly from the population used for calibration). In this paper, we propose a greatly more robust and deployable approach to conformal prediction that relies solely on the observed measurements \( \{y_i\}_{i=1}^M \). %We achieved it using Stein's Unbiased Risk Estimate (SURE) \cite{stein1981estimation} and by  leveraging this calibration strategy, conformal prediction sets can be constructed to reflect uncertainty effectively, even in the absence of ground truth data.

\section{Proposed Method}\label{sec:method}

Our proposed self-supervised conformal prediction method circumvents the need for ground truth data by leveraging Stein's unbiased risk estimate (SURE) \cite{stein1981estimation}. 

We begin by pooling together $M$ exchangeable imaging problems, where each problem involves an unknown image $x^{\star}_i$ and an observation $y_i$ which we consider to be a realisation of the conditional random variable $(Y|X=x^{\star}_i) \sim \mathcal{N}(Ax^{\star}_i,\sigma^2\mathbb{I}_m)$. To specify the non-conformity measure, we  consider $\Sigma(y)=A^\top A$ which is a natural choice for approximation for the error inverse-covariance when $(Y|X=x^{\star}_i) \sim \mathcal{N}(Ax^{\star}_i,\sigma^2\mathbb{I}_m)$, as we expect $\hat{x}(Y)$ to be accurate along the leading eigenvectors of $A^\top A$ and the estimation error to concentrate along weak eigenvectors of $A^\top A$. This leads to the non-conformity measure
\begin{equation} \label{score}
  s(x, y) = \frac{1}{m}\|Ax - A\hat{x}(y)\|_2^2,  
\end{equation}
where we recall that $A$ is assumed full-rank, but potentially very poorly conditioned . We require $A$ to be full rank as otherwise $\Sigma(y)$ is only positive semidefinite, implying that the corresponding prediction set can be unbounded.
 
To calibrate without ground truth data, instead of relying on a sample quantile of $\{s(x_i,y_i)\}_{i=1}^M$, we rely on SURE to provide unbiased estimates of $\{s(x_i,y_i)\}_{i=1}^M$ from the observed measurements $\{y_i\}_{i=1}^M$. We then use those noisy quantile estimates for calibration, at the expense of a small amount of bias. 

More precisely, assuming that the estimator $\hat{x}$ is differentiable almost everywhere, the SURE estimate of \eqref{score} is given by  
\begin{equation}\label{eq:SURE}
 \textrm{SURE}(y) =   \frac{1}{m}\|y - A \hat{x}(y)\|_2^2 -\sigma^2 + \frac{2\sigma^2}{m} \text{div} \left(A \hat{x}(y)\right)\,,  
\end{equation}
where \( \operatorname{div}(\cdot) \) denotes the divergence operator \cite{stein1981estimation}. It is easy to show that $\textrm{SURE}(Y)$ provides an estimate of $s(X, Y)$ that is unbiased \cite{stein1981estimation}. Crucially, when $m = \textrm{dim}(y)$ is large, the estimate provided by SURE is not only unbiased but also often very accurate (see \cite{stein1981estimation,bellec2021second} for a theoretical analysis of the variance of SURE and \cite{6545395} for an empirical analysis in an imaging setting). As a consequence, we expect that the conformal calibration quantiles obtained from SURE will be in close agreement with the true quantiles of $s(X,Y)$, ensuring that the resulting conformal prediction sets nearly maintain their desired coverage properties.

With regards to the evaluation of SURE, for some model-based estimators it is possible to identify a closed-form expression \cite{tibshirani2012degrees}. Otherwise, computing SURE requires a numerical approximation of the divergence term. A common approach is the Monte Carlo SURE (MC-SURE) algorithm \cite{ramani2008monte}, which is estimator-agnostic. We use the so-called Hutchinson's stochastic trace approximation method \cite{9054593}, which is more computationally efficient than MC-SURE and does not require hyper-parameter fine-tuning. More precisely, the divergence \( \text{div}(A\hat{x}(y)) \) is approximated as follows:

\begin{align}
  \text{div}(A\hat{x}(y)) = \text{trace}(\boldsymbol{J}_{h(y)}) &=\mathbb{E}[\tilde{n}^\top \boldsymbol{J}_{h(y)}\tilde{n}]  , \\
   &\approx \frac{1}{K} \sum_{i=1}^{K} \tilde{n}_i^\top \boldsymbol{J}_{h(y)}\tilde{n}_i, \label{eq:MC-SURE_H}  \\
   & \approx \frac{1}{K} \tilde{n}^\top \boldsymbol{J}_{\tilde{n}^ \top  h(y )}, \label{eq:MC-SURE_N} 
\end{align}
where $\{\tilde{n}_i\}_{i=1}^K$ are a $K$ i.i.d. standard normal random vectors, \( \boldsymbol{J}_{h(y)} \) is the Jacobian matrix of the predicted measurements \( h(y) = A\hat{x}(y) \) with respect to \( y \), and \eqref{eq:MC-SURE_H} corresponds to Hutchinson's method, which for computationally efficiency we implement \eqref{eq:MC-SURE_N} using automatic differentiation, as suggested in \cite{9054593}. This allows obtaining accurate SURE estimates in a highly efficient manner, even in very large problems.

Adopting a split-conformal strategy, after computing $\hat{x}(y_i)$ and $\textrm{SURE}(y_i)$, for each $i = \{1,\ldots,M\}$ we construct the $(1-\alpha)$-prediction set \( \hat{C}(y_i) \) as:
\[
\hat{C}(y_i) = \{x \in \mathbb{R}^n : \|Ax - A\hat{x}(y_i)\|_2^2/m \leq \hat{Q}^{(i)}_\alpha\},
\]
where \( \hat{Q}^{(i)}_\alpha \) is the top $\frac{\lceil M(1-\alpha)\rceil}{M-1}$-quantile of the sample $\{\textrm{SURE}(y_j)\}_{j=1}^M$ with the $i$th element, $\textrm{SURE}(y_i)$, removed. Note that while the proposed approach has some bias due to the estimation error introduced by SURE, in our experience the bias is small and arguably significantly smaller than the bias that is likely to be incurred due to distribution shift in deployment. It is also worth mentioning that the estimates $\textrm{SURE}(y_i)$ can be computed in parallel. The proposed method is summarised in Algorithm \ref{alg:one} below.

\begin{algorithm}
\caption{SURE-based conformal prediction}\label{alg:one}

\begin{algorithmic}[1]
\Require{Forward operator $A$, noise variance $\sigma^2$, estimator $\hat{x}$, measurement $y$, samples $\{y_i\}_{i = 1}^M$, precision level $1-\alpha \in (0,1) . $}  
\Statex
\For{$i \gets 1$ to $M$}                    
    \State {$S_i \gets \textrm{SURE}(y_i)$ using \eqref{eq:SURE} and \eqref{eq:MC-SURE_N}}
    % \If{$s_{\text{mul}}$}
    %     \State $s_i \gets \frac{\|Ax_i-A\hat{x}(y_i)\|_2^2}{S_i}$
    % \ElsIf{$s_{\text{add}}$}
    %     \State $s_i \gets \|Ax_i-A\hat{x}(y_i)\|_2^2 - S_i$
    % \EndIf
\EndFor

\State {$\hat{Q}_\alpha$ $\gets$ top $\frac{\lceil(M+1)(1-\alpha)\rceil}{M}$-quantile of $\{S_i\}_{i = 1}^M$}
\State $m  \gets dim(y)$

% \State {$S \gets \textrm{SURE}(y)$ using \eqref{eq:SURE} and \eqref{eq:MC-SURE_2}}

% \If{$s_{\text{mul}}$}
\State {$\hat{C}(y)$ $\gets$ $\{x\in \mathbb{R}^n\,|\, \|Ax-A\hat{x}(y)\|_2^2 / m\leq  \hat{Q}_\alpha\}$}
%% \ElsIf{$s_{\text{add}}$}
% \State {$\hat{C}(y)$ $\gets$ $\{x\in \mathbb{R}^n\,|\, \|Ax-A\hat{x}(y_{\text{new}})\|_2^2 \leq S + \hat{Q}_\alpha\}$}
% \EndIf
\Statex 
\Ensure{Prediction set $\hat{C}(y)$}
\end{algorithmic}
\end{algorithm}

\section{Experiments}\label{sec:experiments}
We demonstrate the proposed self-supervised conformal prediction approach by applying it to two image restoration problems: {image denoising} and {image deblurring}. To showcase the versatility of the method, for image denoising we construct conformal prediction sets by using a learning-based image restoration technique trained in a self-supervised end-to-end manner, whereas for image deblurring we use the model-driven technique. %Fig. \ref{fig:reconstruction} for examples from these two problems.
For each experiment, we implement our method by using Algorithm \ref{alg:one} to compute conformal prediction sets. We consider a fine grid of values for the confidence parameter \( \alpha \in (0,1) \), ranging from $0\%$ to $100\%$. The corresponding prediction sets should cover the solution space with a probability of approximately \( 1-\alpha \). We evaluate the accuracy of these prediction sets by calculating the empirical coverage probabilities on a test set. Specifically, we compute the proportion of test images that lie within the conformal prediction sets for various confidence levels. In all experiments, the calibration sample size \( M \) is chosen to be sufficiently large to ensure that the variability of the sample quantiles caused by the finite-sample correction is negligible. Furthermore, for comparison, in each experiment we also report results by supervised conformal prediction (i.e., using ground truth data for calibration, rather than SURE). %, as well as results from the equivariant bootstrapping technique proposed recently in \cite{tachella2023equivariant}. 
The experiments are implemented using the Deep Inverse library\footnote{https://deepinv.github.io/deepinv/} for imaging with deep learning using PyTorch.

%Furthermore, 

%we provide comparisons with a supervised conformal prediction to the equivariant bootstrap, which our approach greatly outperforms on these image restoration problems. Additionally, we compare the self-supervised conformal prediction method with its supervised counterpart, which achieves the most accurate uncertainty quantification. Remarkably, the results show that the self-supervised approach is very close to, and in some cases (deblurring See Fig. \ref{fig:deblurring})  almost identical to, its supervised version, highlighting its effectiveness and reliability. The experiments are conducted using the Deep Inverse library\footnote{https://deepinv.github.io/deepinv/} for imaging with deep learning using PyTorch. 

\begin{figure}[h!]
    \centering
    \includegraphics[width=1\linewidth]{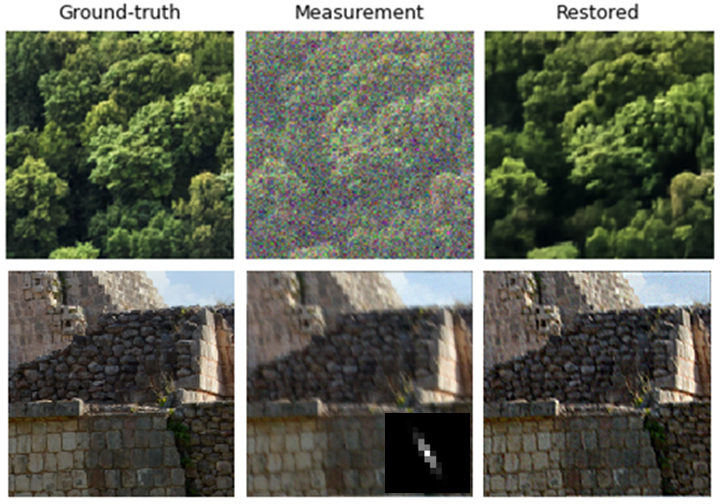}
    \caption{\textbf{Image reconstruction results for various imaging problems}. Top:  Gaussian noise on DIV2K.  Bottom:
Noisy Gaussian blur on DIV2K }
    \label{fig:reconstruction}
\end{figure}

%, and the results highlight the ability of the proposed method to provide accurate uncertainty quantification across a range of restoration tasks.

% We now illustrate the proposed conformal prediction approach by applying it to three image restoration problems: image denoising, image deblurring and computed tomography. To show the versatility of the method, we consider the construction of conformal prediction sets for both model-driven and data-driven image restoration techniques. We conduct the following image restoration experiments using the deep inverse library \cite{Tachella_DeepInverse_A_deep_2023} :

\subsection{Image Denoising}
For the image denoising experiment, we consider colour images of size $128 \times 128$ pixels obtained by cropping images from the DIV2K dataset \cite{Agustsson_2017_CVPR_Workshops}, which we artificially degrade by adding white Gaussian noise with a standard deviation of \( \sigma = 0.1 \). As image restoration method, we use a DRUNet model \cite{zhang2021plug} trained in a self-supervised manner by using the SURE loss \cite{tachella2024unsure}. The training data consists of $900$ noisy measurements, we do not use any form of ground truth for training or for conformal prediction. See the top row of Fig. \ref{fig:reconstruction} for an example of a clean image, its noisy measurement, and the estimated reconstruction. We then use these same noisy images to compute our proposed self-supervised conformal prediction sets, and assess their accuracy empirically by using $200$ measurement-truth pairs from the test dataset. The results are reported in Fig. \ref{fig:denoising} below, together with the results obtained by using the equivalent supervised conformal prediction approach that relies on ground truth data for calibration. We observe that our method delivers prediction sets that are remarkably accurate and in close agreement with the results obtained by using supervised conformal prediction, demonstrating that the bias stemming from using SURE instead of ground truth data is negligible in this case. For completeness, Fig. \ref{fig:histogram} (left) below shows the empirical distribution of the non-similarity function $s(x,y)$ for the supervised conformal prediction based on the MSE, and the proposed self-supervised conformal prediction based on a SURE estimate of the MSE. Again, we observe close agreement between these distributions, with the SURE distribution being slightly more spread due to the random error inherent to SURE. %Moreover, for completeness we also report the results obtained by equivariant bootstrapping \cite{tachella2023equivariant}, which is significantly less accurate for this problem\footnote{The bootstrap is implemented by using rotation and two-dimensional shifts. Rotations are sampled from a Gaussian distribution with zero mean and a standard deviation of \( \sigma_{\theta} \) (denoising: \( \sigma_{\theta} = 125 \), deblurring: \( \sigma_{\theta} = 200 \)), while horizontal and vertical shifts are sampled from a uniform distribution on \( [-\Delta t, \Delta t] \) pixels (denoising: \( \Delta t = 250 \)}. The lack of accuracy of equivariant bootstrapping stems 

\begin{figure}[h!]
    \centering
    \includegraphics[width=0.75\textwidth]{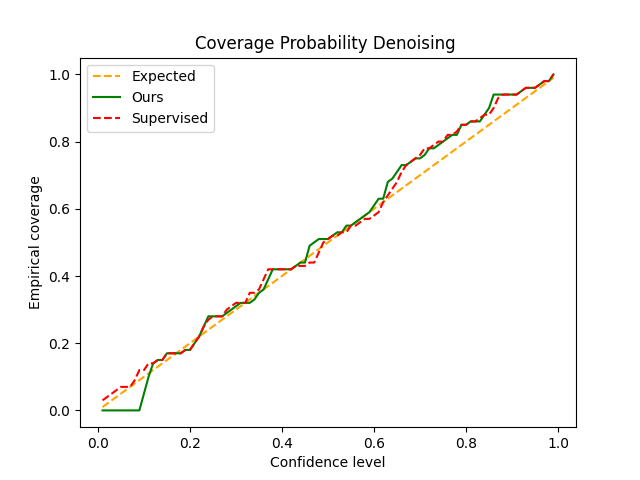}
    \caption{\textbf{Image denoising experiment}: desired confidence level vs empirical coverage. Both supervised and the proposed self-supervised conformal prediction methods deliver prediction sets with near perfect coverage.}
    \label{fig:denoising}
\end{figure}

\subsection{Image Deblurring}
We now consider a non-blind image deblurring experiment with colour images of size $256 \times 256$ pixels, derived from the DIV2K dataset \cite{Agustsson_2017_CVPR_Workshops} and artificially degraded with a diagonal Gaussian blur of major bandwidth $\varsigma_0 = 2$, $\varsigma_1 = 0.3$ along the minor axis, and an inclination of ${\pi}/{6}$ degrees, as well as additive white Gaussian noise  with a standard deviation \( \sigma = 0.01 \). As image restoration technique, we use the recently proposed model-based Polyblur technique \cite{delbracio2021polyblurremovingmildblur}, a highly efficient restoration method for removing mild blur from natural images based on a truncated polynomial approximation of the inverse of the blur kernel. See the bottom row of Fig. \ref{fig:reconstruction} for an example of a clean image, its noisy measurement, and the estimated reconstruction. 

\begin{figure}[h!]
    \centering
    \begin{subfigure}[h!]{0.49\textwidth}
        \includegraphics[width=\textwidth]{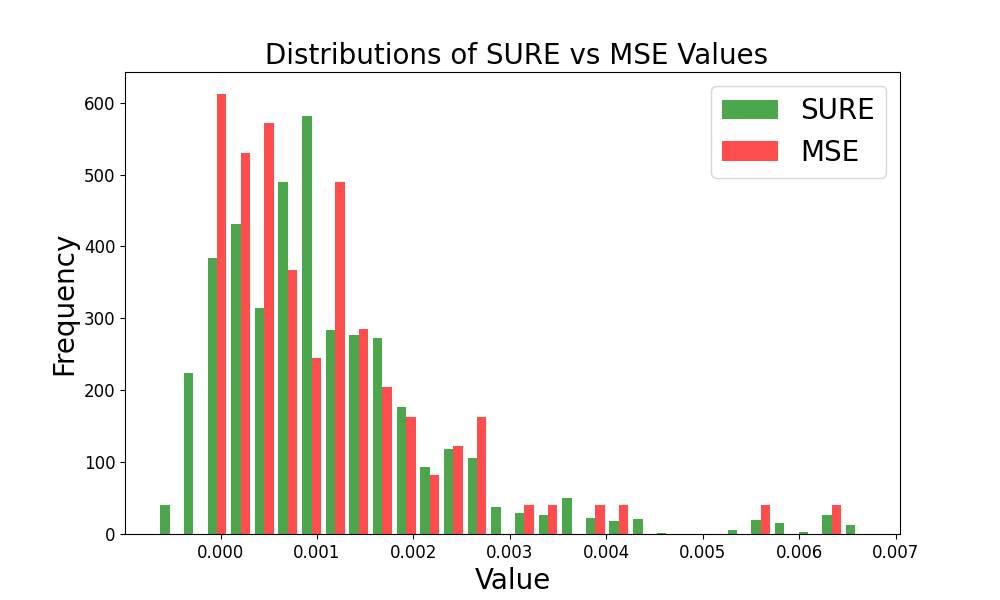}
        \caption{Image denoising}
        \label{subfig:histogramdenoising}
    \end{subfigure}
    %\hfill % this command adds a little space between the images
    \begin{subfigure}[h!]{0.49\textwidth}
        \includegraphics[width=\textwidth]{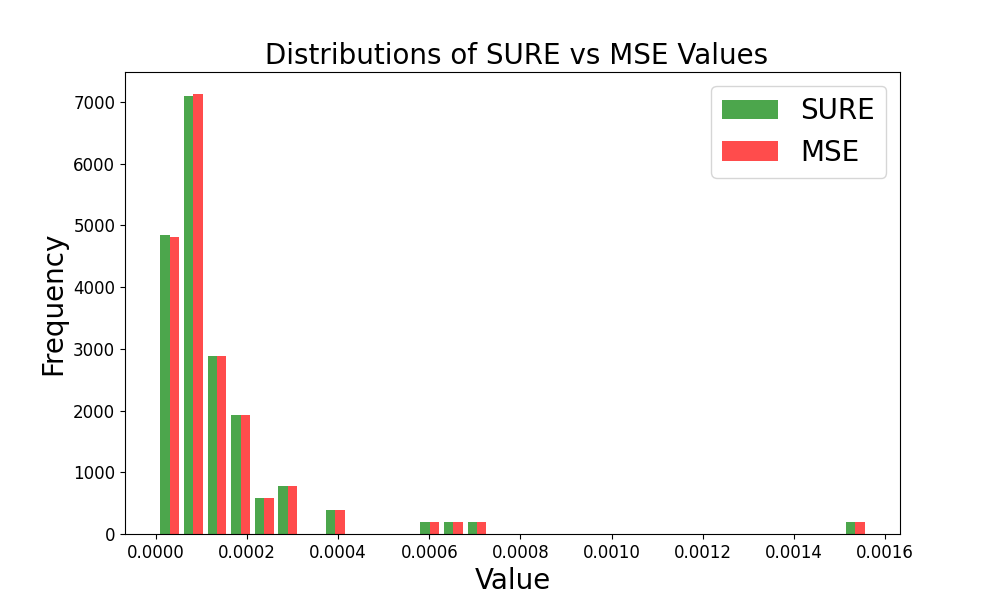}
        \caption{Image deblurring}
        \label{subfig:histogramdeblurring}
    \end{subfigure}
    \caption{\textbf{Calibration histograms} (empirical distribution of the non-similarity function $s(x,y)$) for the supervised case (MSE) and the self-supervised case based on a SURE estimate of the MSE, for the denoising and deblurring experiments.}
        \label{fig:histogram}
\end{figure}

We use $900$ blurred and noisy images to compute our proposed self-supervised conformal prediction sets, and assess their accuracy empirically by using $200$ measurement-truth pairs from the test dataset. The results are reported in Fig. \ref{fig:deblurring} below, together with the results obtained by using the equivalent supervised conformal prediction approach that relies on ground truth data for calibration. Again, we observe that our method delivers prediction sets that are accurate and remarkably close to the results obtained by using supervised conformal prediction, demonstrating that the bias stemming from using SURE instead of ground truth data is again negligible in this case. For completeness, Fig. \ref{fig:histogram} (right) below shows the empirical distribution of the non-similarity function $s(x,y)$ for the supervised conformal prediction based on the MSE, and the proposed self-supervised conformal prediction based on a SURE estimate of the MSE. Again, we observe close agreement between these distributions.

\begin{figure}[h!]
    \centering
    \includegraphics[width=0.75\textwidth]{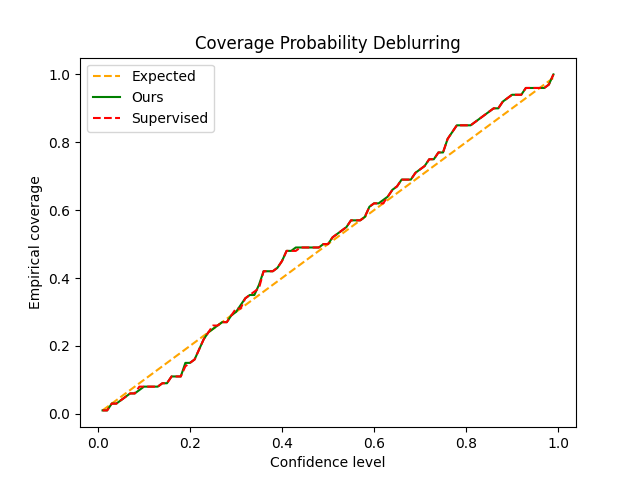}
  \caption{\textbf{Image deblurring experiment}: desired confidence level vs empirical coverage. Both supervised and the proposed self-supervised conformal prediction methods deliver prediction sets with near perfect coverage.}
    \label{fig:deblurring}
\end{figure}

\section{Discussion and Conclusion}\label{sec:discussion}
This paper presented a self-supervised approach for constructing conformal prediction regions for linear imaging inverse problems that are ill-posed. Unlike previous conformal prediction methods, the proposed approach does not require any form of ground truth data. This is achieved by leveraging Stein's unbiased risk estimator and by pooling together a group of exchangeable imaging problems. This allows delivering conformal prediction sets in situations where there is no ground truth data available for calibration, and provides robustness to distribution shift. Additionally, the proposed framework is estimator-agnostic, as it uses a Monte Carlo implementation of SURE that does not require any explicit knowledge of the image restoration algorithm used. This flexibility allows the framework to be straightforwardly applied to model-driven as well as data-driven image restoration techniques, including self-supervised data-driven techniques that are also trained directly from the measurement data. Moreover, the method is computationally efficient as computations can be performed in parallel. We demonstrated the effectiveness of the proposed method through image denoising and deblurring experiments, where we observed that our method delivers extremely accurate prediction sets. 

Future work will explore the generalization of the proposed approach to other noise models, such as Poisson and Poisson-Gaussian, as well as to problems in which the parameters of the noise model are unknown \cite{tachella2024unsure}. Another important perspective for future work is to extend this approach to problems in which the forward $A$ is not full rank, for example by leveraging equivariance properties. In particular, it would be interesting to study the integration of our proposed method and the equivariant bootstrap \cite{tachella2023equivariant}.

\bibliographystyle{IEEEtranN}
\bibliography{references}

\end{document}